%% file: main.tex
\documentclass{article}
\usepackage[dblblindworkshop,final]{neurips2026}
\workshoptitle{World Models for High-Stakes Health}

\usepackage[switch]{lineno}
\usepackage[utf8]{inputenc} 
\usepackage[T1]{fontenc}    
\usepackage{hyperref}       
\usepackage{url}            
\usepackage{booktabs}       
\usepackage{amsfonts}       
\usepackage{nicefrac}       
\usepackage{microtype}      
\usepackage{xcolor}         
\usepackage{graphicx}
\usepackage{amsmath}
\usepackage{amssymb}
\usepackage{makecell}
\usepackage{float}
\usepackage{placeins}
\usepackage{tikz}
\usepackage{pgfplots} 
\usepackage{booktabs} 
\usetikzlibrary{shapes.geometric, arrows.meta, positioning, calc, matrix}
\pgfplotsset{compat=1.18}
\setcitestyle{numbers,square,comma}
\usepackage{wrapfig}

\title{Are Joint-Embedding Predictive Objectives Effective for Longitudinal EHRs?}

\author{%
  Ashvin Gupta$^{1}$ \quad Brendan Delaney$^{1}$ \quad Alessandra Russo$^{1}$ \\[0.5em]
  $^{1}$Imperial College London \\
}

\begin{document}

\maketitle

\input{text/abstract}

\input{text/introduction}
\input{text/methods}
\input{text/experiments_results}
\input{text/discussion}

\bibliography{references} 
\bibliographystyle{unsrtnat}

\input{text/appendix}

\end{document}

%% file: text/abstract.tex
\begin{abstract}
Masked language modelling (MLM) is the dominant pretraining objective for transformer encoders applied to longitudinal electronic health records (EHRs). However, reconstructing individual events may overemphasise noisy or weakly informative observations and does not explicitly encourage learning higher-level patient states. Joint-Embedding Predictive Architectures (JEPAs) instead learn by predicting representations in latent space and have shown strong performance in a variety of domains. 
In this paper, we investigate to what extent JEPA can be adapted to EHR sequence representation.
Our empirical investigation using MIMIC-IV shows that although all three of our JEPA formulations train stably without representational collapse, they consistently underperform standard MLM.
We find that target size, not target heterogeneity, increases prediction error, yet reducing the target size does not recover downstream performance.
This suggests that stable latent  prediction alone is insufficient for competitive EHR representation learning and highlights the challenge of defining clinically meaningful latent targets for EHR.

\end{abstract}

%% file: text/introduction.tex
\section{Introduction}

Electronic Health Records (EHRs) are longitudinal records of patient-healthcare interactions. Recent work has increasingly focused on EHR foundation models, which aim to learn general-purpose patient representations through self-supervised learning and can be adapted to a range of downstream clinical prediction tasks \citep{Steinberg2020LanguageData,Renc2024ZeroTransformer, Shmatko2025LearningTransformers, Rajamohan2025FoundationScale}. 
Downstream performance therefore hinges on the quality of these representations.

Due to the sequential nature of EHRs, transformer encoders have become a common backbone for EHR representation learning.  
A dominant objective for pretraining has been BERT-style masked language modelling (MLM), in which 
the model is trained to reconstruct masked medical events from their surrounding context \citep{Li2019BEHRT:Records, Li2021Hi-BEHRT:Records, Rasmy2021Med-BERT:Prediction, Pang2021CEHR-BERT:Tasks}. However, exact token reconstruction may not provide the ideal learning signal for longitudinal clinical data. EHR sequences contain routine, repeated, administrative, and partially unpredictable events, yet MLM treats reconstruction of each masked token as equally important. This may encourage representations that capture the identity of local events rather than the true patient state.

In contrast, clinicians use their medical knowledge to abstract away irrelevant detail and focus on key events, trends, and interactions to form a higher-level representation of the patient state. This motivation is closely aligned with Joint-Embedding Predictive Architectures (JEPAs), which provide an alternative form of self-supervision \citep{LeCun2022AIntelligence, Assran2023Self-SupervisedArchitecture}. Rather than reconstructing observations in the input space, JEPA trains a predictor to estimate latent representations of unobserved targets from an available context. This approach has shown strong representation-learning performance in vision \citep{Assran2023Self-SupervisedArchitecture, Assran2025V-JEPAPlanning}, world modelling \citep{Maes2026LeWorldModel:Pixels} and language \citep{Huang2025LLM-JEPA:Architectures}.
Latent prediction is intended to produce semantically meaningful features and 
avoid unnecessary reconstruction of unpredictable low-level detail. This is particularly appealing for EHRs,
where multiple observed event sequences may correspond to similar underlying clinical states and trajectories.

However, defining a meaningful and predictable JEPA target in longitudinal EHRs is non-trivial. Unlike image patches, EHR events are irregularly sampled, vary substantially in density, and already represent semantically meaningful concepts. Temporal proximity therefore does not necessarily imply semantic similarity, and future clinical events may depend not only on patient state but also on healthcare decisions.

In this work, we ask whether the I-JEPA-style latent prediction framework \citep{Assran2023Self-SupervisedArchitecture} can be transferred effectively to longitudinal EHR sequences. We evaluate multiple target-construction strategies, including 
\textit{span} masking, \textit{future} masking and \textit{random} masking of events. 
We investigate (1) whether these JEPA formulations can be trained stably without representational collapse, (2) how the downstream performance compares, and (3) which objective and training characteristics may explain the observed differences.


%% file: text/methods.tex
\section{Methodology}

\subsection{Medical Event Representation}

\begin{wrapfigure}{r}{0.48\linewidth}
    \centering
    \includegraphics[width=\linewidth]{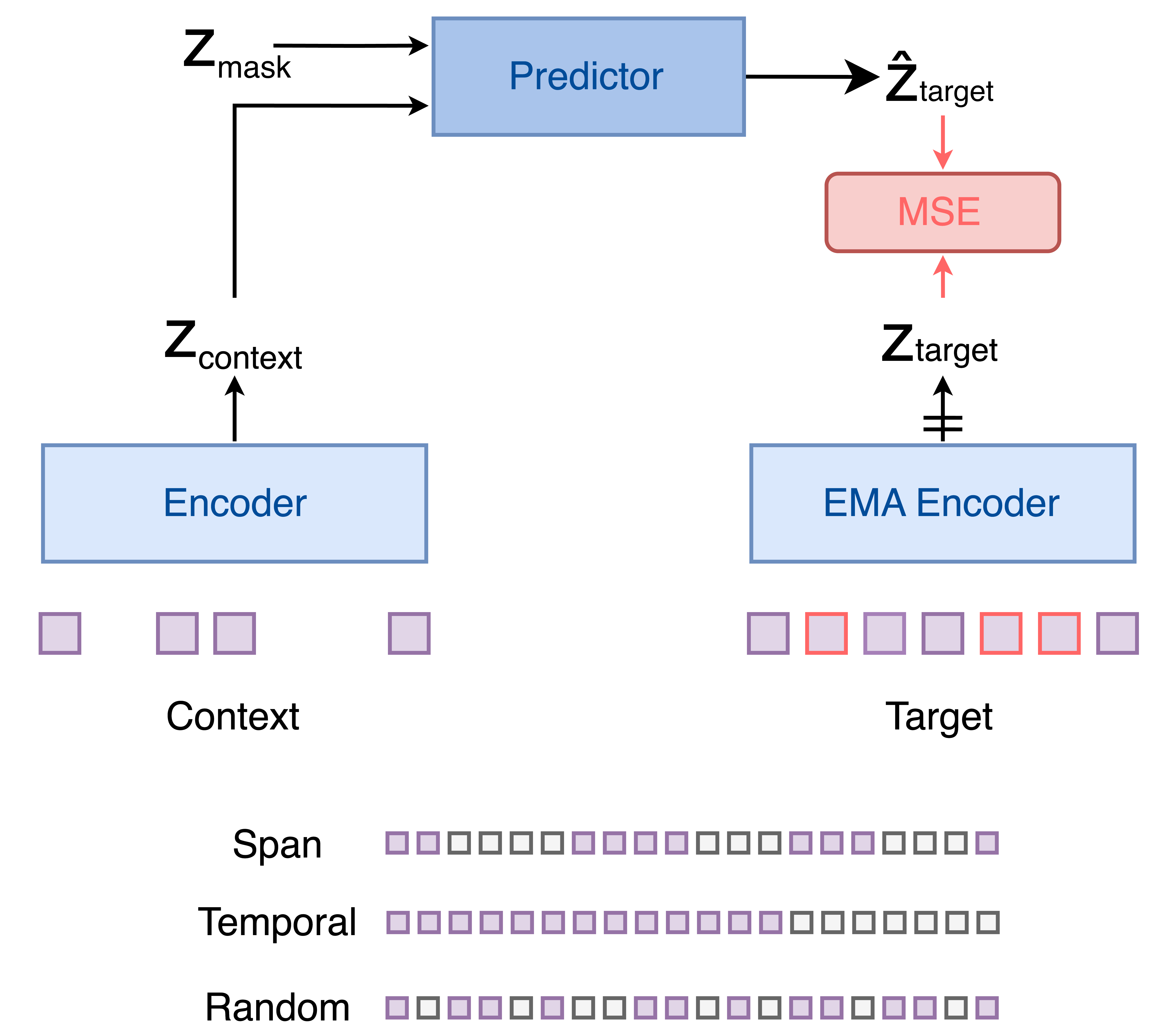}
    \caption{The context sequence is encoded by the online encoder, while the complete target sequence is processed by an exponential-moving-average (EMA) target encoder with gradients stopped through the target branch. A predictor combines the context representations with learnable mask embeddings to estimate target latents, which are matched to the EMA target representations using mean squared error. We evaluate three target constructions: contiguous span masking, prospective prediction of future events, and random masking of individual events.}
    \label{fig:MLM_JEPA}
\end{wrapfigure}

A patient trajectory is represented as a time-ordered sequence of clinical events, $X=(e_1,e_2,\ldots,e_N)$, where each event is a triplet $e_i=(c_i,v_i,t_i)$ comprising a discrete medical code $c_i\in\mathcal{V}$, an optional continuous value $v_i\in\mathbb{R}$, and timestamp $t_i$. Patient age $a_i$ is derived from $t_i$ and date of birth. Representing time and continuous values is an important design choice in EHR modelling \citep{Attrach2025RethinkingMore, Montgomery2026HowRecords, Guo2026TokenizationModels}. Rather than discretising continuous recordings into tokens, we retain them directly in the event representation using triplet embeddings \citep{Attrach2025RethinkingMore}. The input embedding for event $e_i$ is

\begin{equation}
\mathrm{h_i} =
{\mathrm{E_{code}}}(c_i)
+
\mathrm{Proj_{val}}(v_i)\cdot\mathbb{I}_{[v_i \neq \varnothing]}
+
\mathrm{Proj_{age}}(a_i),
\end{equation}

where $\mathrm{E_{code}}$ is a learned code embedding and $\mathbb{I}$ indicates value presence. Sequence order is encoded using Rotary Position Embeddings (RoPE). All models use the same transformer backbone with 6 layers, 8 attention heads, hidden dimension 768, and maximum context length $N=1024$, ensuring that downstream differences reflect the pretraining objective rather than model capacity.

\subsection{Objectives}
JEPA performance depends strongly on context--target design \citep{Huang2025LLM-JEPA:Architectures}, and as such, we evaluate three EHR target constructions reflecting different assumptions about what should be predictable.

Our first objective, \emph{Span JEPA}, follows the I-JEPA paradigm as directly as possible \citep{Assran2023Self-SupervisedArchitecture}. 
We treat the EHR as a one-dimensional sequence of $N$ clinical events. The complete uncorrupted sequence is passed through the exponential moving average (EMA) target encoder to obtain event-level representations $[
\mathbf{z}_y =
{\mathbf{z}_{y_1},\ldots,\mathbf{z}_{y_N}}.
]$. 
We sample $M$ contiguous target spans from these representations. The (i)-th target span is defined by the set of event positions $B_i$, giving the corresponding target representations ${\mathbf{z}_{y_j}},{j \in B_i}.$
The target events are removed from the input leaving the remaining events as context. The context encoder produces a set of event representations, which are passed to a shallow two-layer transformer predictor together with learnable mask tokens containing positional information for the target locations. As in I-JEPA, the predictor is applied separately for each target span, producing ${\hat{\mathbf{z}}_{y_j}},{j \in B_i}.$
The prediction objective is the mean squared distance between the predicted and target representations across the $M$ target spans:



\begin{equation}
\mathcal{L}_{\mathrm{JEPA}}
=
\frac{1}{M}
\sum_{i=1}^{M}
\frac{1}{|B_i|}
\sum_{j \in B_i}
\left\|
\hat{\mathbf{z}}^{(i)}_{y_j}
-
\mathbf{z}_{y_j}
\right\|_2^2.
\end{equation}


While span masking provides the closest analogue to I-JEPA, EHRs are inherently temporal, and clinically meaningful information may instead lie in how the patient evolves over time. We therefore consider a prospective formulation, \emph{Future JEPA}, motivated by temporally predictive JEPA and world-model approaches \citep{Assran2025V-JEPAPlanning}. All events observed up to a time point $T$ form the context, while events satisfying $[T < t_j \leq T + 24,\mathrm{h}]$ form a single future target region $B_1$. This is equivalent to setting $M=1$ in the JEPA objective above. The architecture and loss are otherwise unchanged. 

Finally, we evaluate a random-mask JEPA formulation, \emph{Random JEPA}, that uses the same random sampling strategy as MLM, but optimises the JEPA objective with $M=1$. This provides a controlled comparison with MLM under the same corruption scheme, differing only in whether the model predicts discrete medical codes or continuous target representations.



Together, these objectives test three increasingly different assumptions about predictable structure in EHRs: (1) whether latent clinical information can be recovered from neighbouring events, (2) whether it can be predicted prospectively from earlier patient history, and (3) whether latent reconstruction offers an advantage over token reconstruction when the corruption pattern itself is held fixed. 


For all JEPA experiments, we use an EMA target encoder with stop-gradient, rather than the shared-encoder SIGReg alternative \citep{Balestriero2025LeJEPA:Heuristics}. In preliminary experiments, SIGReg was unstable and collapsed, with substantially reduced target-embedding variance. The EMA formulation remained stable and non-collapsed, allowing us to assess the JEPA objective without conflating downstream performance with optimisation failure. Standard MLM is retained as the primary baseline.

\begin{table*}[t]
\centering
\caption{Performance (AUROC and AUPRC) on downstream clinical prediction tasks on MIMIC-IV for different pretraining objectives and masking strategies.}
\label{tab:main_results}
\resizebox{\textwidth}{!}{
\begin{tabular}{llcccccccc}
\toprule
& & \multicolumn{2}{c}{ICU Mortality} & \multicolumn{2}{c}{ICU Readmission} & \multicolumn{2}{c}{Hospital Mortality} & \multicolumn{2}{c}{Hospital Readmission} \\
\cmidrule(lr){3-4}
\cmidrule(lr){5-6}
\cmidrule(lr){7-8}
\cmidrule(lr){9-10}
Loss & Masking &
AUROC & AUPRC &
AUROC & AUPRC &
AUROC & AUPRC &
AUROC & AUPRC \\
\midrule

MLM & Random & \textbf{0.774} & \textbf{0.241} & \textbf{0.695} & \textbf{0.269} & \textbf{0.878} & \textbf{0.266} & \textbf{0.720} & \textbf{0.416}\\
JEPA & Span & 0.715 & 0.172 & 0.642 & 0.219 & 0.845 & 0.210 & 0.700 & 0.401\\ 
JEPA & Future & 0.757 & 0.216 & 0.681 & 0.259 & 0.870 & 0.249 & 0.708 & 0.409 \\ 
JEPA & Random & 0.715 & 0.179 & 0.638 & 0.215 & 0.841 & 0.214 & 0.691 & 0.390\\ 
\bottomrule
\end{tabular}}
\end{table*}

%% file: text/experiments_results.tex
\section{Experiments and Results}

Each objective is trained for 50 epochs on the full MIMIC-IV training cohort using the same optimisation setup. Our downstream tasks include ICU mortality, ICU readmission, hospital mortality, and hospital readmission using the same 80:10:10 patient-level split as pretraining. Representation quality is assessed using frozen-encoder linear probes with mean pooling, using the checkpoint with the lowest validation pretraining loss, reporting AUROC and AUPRC. The full hyperparameter sweep is provided in Appendix~\ref{Hyperparameters}.

MLM outperformed all JEPA formulations across downstream tasks, with \emph{Future JEPA} performing closest to MLM, followed by the two others (Table~\ref{tab:main_results}). To determine whether this performance gap was caused by representational collapse, we additionally tracked effective rank and feature standard deviation of the target-encoder embeddings throughout pretraining \citep{Garrido2023RankMe:Rank}. Both remained stable while training and validation prediction losses decreased (Appendix \ref{Training_dynamics}, Figure \ref{fig:training_curves}), indicating that the JEPA models converged to non-collapsed solutions despite learning representations that were less useful for downstream clinical prediction.

\begin{wrapfigure}{r}{0.48\linewidth}
    \centering
    \includegraphics[width=\linewidth]{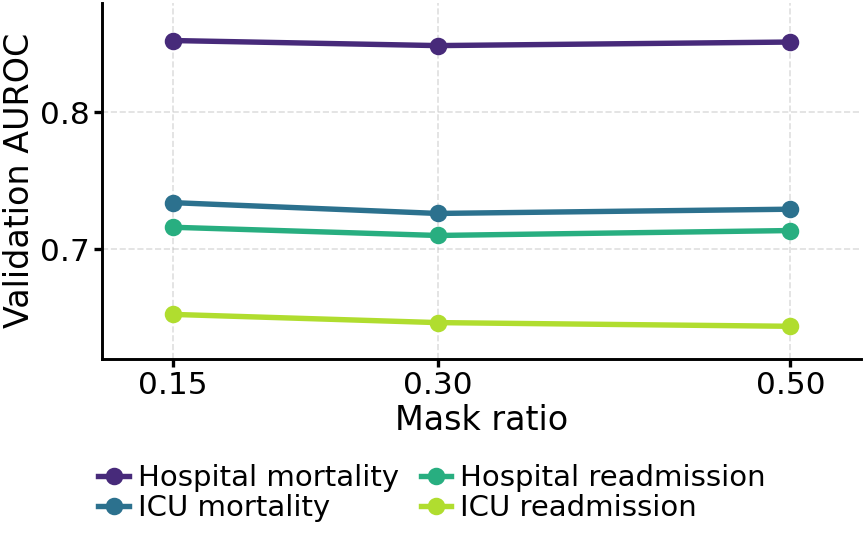}
    \caption{Sensitivity of Span-JEPA downstream performance to target masking ratio across the four downstream clinical prediction tasks.}
    \label{fig:mask_sweep}
\end{wrapfigure}

We further investigated which properties of an EHR target were associated with latent-prediction difficulty.
We performed a post-hoc multivariable regression on per-target prediction error using target size, unique-code fraction, event-type diversity, target--context code overlap, and code rarity as covariates (Appendix~\ref{regression}). Target size, a key design parameter in I-JEPA \citep{Assran2023Self-SupervisedArchitecture},
was the only factor consistently positively associated with prediction error across all three formulations. In contrast, unique-code fraction was negatively associated with prediction error, while event-type diversity had comparatively small effects. To investigate the effect of target size, we varied the masking ratio to alter target size for \emph{Span JEPA}. Although smaller targets reduced prediction difficulty, this produced no consistent improvement in downstream performance (Figure~\ref{fig:mask_sweep}).

%% file: text/discussion.tex
\section{Discussion}

Direct JEPA-style latent prediction can be optimised successfully on longitudinal EHRs without representational collapse, yet the resulting representations remain weaker than MLM on the evaluated downstream tasks. The fact that reducing target size also fails to improve downstream performance suggests that the limitation is not simply optimisation difficulty, but may instead lie in how the prediction target is defined.

A likely explanation is a mismatch in the semantic unit of prediction. In I-JEPA, image patches are relatively low-level observations, and predicting larger regions can encourage abstraction toward higher-level structure \citep{Assran2023Self-SupervisedArchitecture}. 
However, EHR tokens already have a very specific semantic meaning associated to them.
Predicting masked event-level latents requires reconstructing the semantic meaning of each event, rather than definitely encouraging compression into a coherent patient-state representation. In this setting, increasing the amount of masked information does not necessarily increase semantic abstraction in the same way that predicting larger image regions can. By contrast, MLM retains an explicit token-level objective that directly preserves event identity, which itself is clinically informative. This may also explain why recent JEPA-style EHR work retains explicit token-level supervision alongside latent prediction \citep{Adam2026TheEHR}.

The stronger performance of \emph{Future JEPA} suggests that temporal prediction may provide a more appropriate inductive bias than reconstructing arbitrary internal targets. However, future EHR observations are only partially determined by latent patient state and are also shaped by clinical decisions. The target may therefore contain information about how care is delivered as well as how the patient evolves. Irregular sampling further complicates target construction, since a fixed number of events can span very different clinical intervals, while a fixed temporal window can contain widely varying amounts of information. Consequently, neither token contiguity nor temporal proximity necessarily corresponds to a coherent patient-state target.

These observations suggest that an EHR-specific JEPA may need to operate at a higher semantic level. Rather than predicting one latent representation per recorded event, groups of events could first be aggregated into a smaller number of patient-state representations, with the predictive objective modelling transitions between these states over time. Such targets could better separate persistent clinical state from individual observations, potentially matching the abstraction motivating JEPA more closely. 

Overall, our findings suggest that the central challenge in transferring JEPA to longitudinal EHRs is not simply preventing collapse or tuning target difficulty, but defining a prediction unit that corresponds to clinically meaningful patient state. In the formulations studied here, successful latent prediction alone is insufficient to outperform token-level reconstruction.

%% file: text/appendix.tex
\onecolumn
\section{Appendix}

\subsection{Responsible Use}
This study evaluates JEPA techniques for retrospective prediction from longitudinal EHRs and does not establish suitability for clinical deployment. Experiments are limited to MIMIC-IV, and performance may not generalise across institutions, patient populations, or care settings. Generative predictions may exhibit uncertainty, miscalibration, and clinically consequential errors. Before any clinical application, models would require external validation, subgroup fairness assessment, calibration analysis, and prospective evaluation under appropriate clinical oversight. Outputs should not be used to guide patient care without such validation and human supervision and as such, this work should be used for research purposes only.

\subsection{Acknowledgements}
This work is supported by UK Research and Innovation (UKRI AI Centre for Doctoral Training in Digital Healthcare grant number EP/Y030974/1), NIHR Imperial Biomedical Research Centre and the NIHR Invention for Innovation grant ``Artificial Intelligence to support cancer early diagnosis in general practice" (NIHR207533).

\subsection{Training Dynamics}
\label{Training_dynamics}
\begin{figure}[h]
    \centering
    \includegraphics[width=0.8\textwidth]{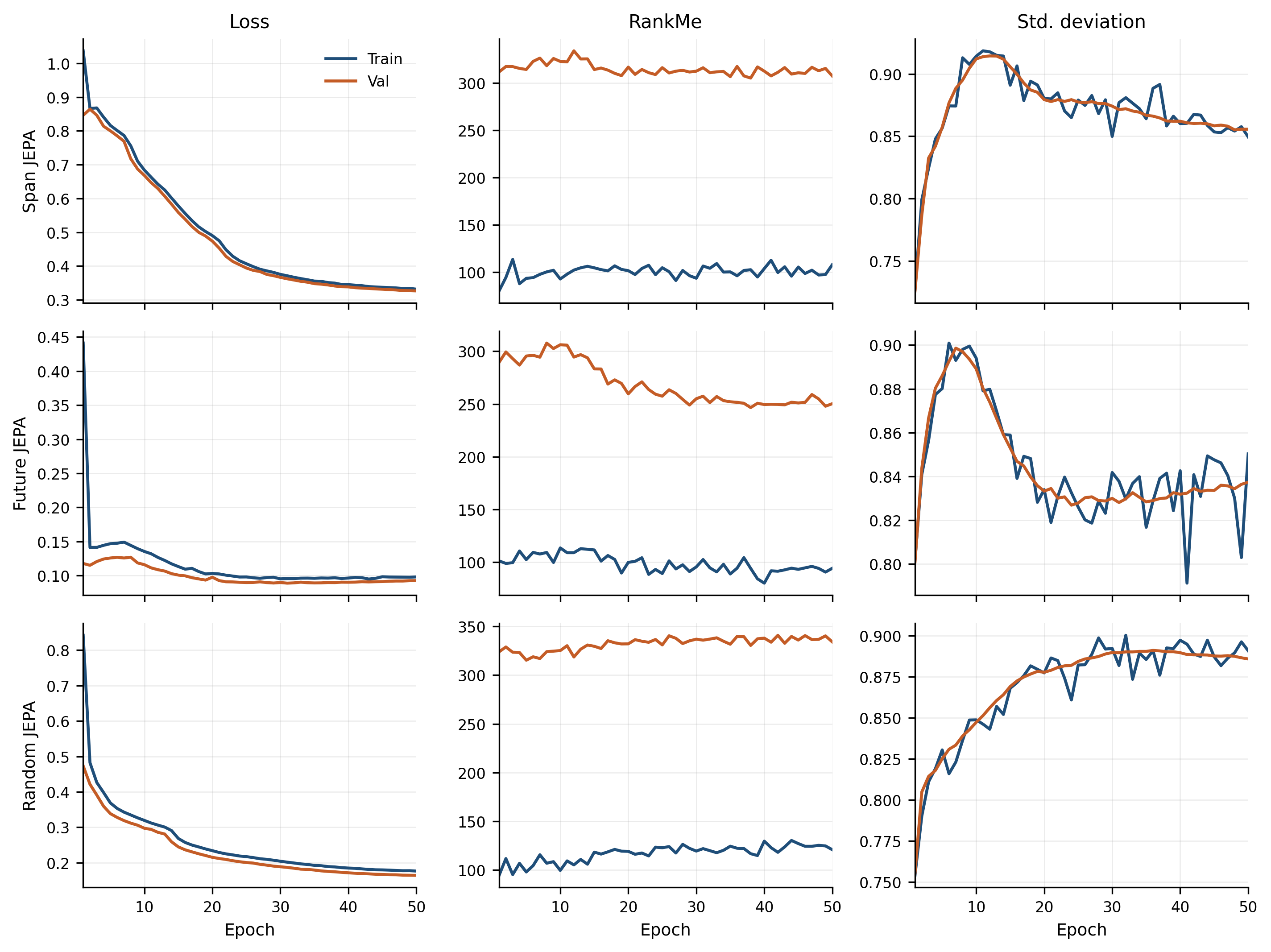}
    \caption{Training dynamics for the three JEPA formulations over 50 epochs. Rows correspond to Span, Future, and Random JEPA, while columns show prediction loss, effective rank (RankMe) of the target-encoder embeddings, and feature standard deviation. Training and validation prediction losses decrease throughout pretraining, while effective rank and feature variance remain non-degenerate, providing evidence that the models do not undergo representational collapse.}
    \label{fig:training_curves}
\end{figure}

\subsection{Regression}
\label{regression}

To investigate which properties of an EHR target are associated with latent-prediction difficulty, we perform a post-hoc regression analysis on the validation set for each JEPA formulation. For every target, we record the JEPA mean squared prediction error (MSE) together with structural characteristics of the target and its relationship to the available context. Because the per-target MSE distribution is strongly right-skewed, we use $\log(1+\mathrm{MSE})$ as the regression outcome. All predictor variables are standardised to zero mean and unit variance before fitting, allowing their coefficients to be compared on a common scale. We fit ordinary least-squares regression models using HC3 heteroskedasticity-robust standard errors and report 95\% confidence intervals.

The regression covariates include target size, unique-code fraction, event-type diversity, target--context code overlap, and code rarity. Target size is defined as the number of events being predicted. Unique-code fraction is the number of unique medical codes in the target divided by the total number of target events. Event-type diversity is measured using normalised entropy over the event types present in the target. Target--context code overlap is the proportion of unique target codes that are also observed in the context, while code rarity is calculated as the mean negative log-frequency of the target codes using frequencies estimated from the pretraining set.
We assess multicollinearity using pairwise correlations and variance inflation factors (VIFs). 
The regression can be written schematically as
\begin{equation}
\log(1+\mathrm{MSE}_i)
=
\beta_0
+
\beta_1 \log(1+N_{\mathrm{target},i})
+
\beta_2 U_i
+
\beta_3 H_i
+
\beta_4 O_i
+
\beta_5 R_i
+
\epsilon_i,
\end{equation}

where $N_{\mathrm{target},i}$ is target size, $U_i$ is unique-code fraction, $H_i$ is event-type diversity, $O_i$ is target--context code overlap, $R_i$ is code rarity.

\begin{table*}[t]
\centering
\caption{Detailed regression results for JEPA prediction error. Coefficients are standardised; confidence intervals and $p$-values use HC3 robust standard errors.}
\label{tab:regression_appendix}
\resizebox{\textwidth}{!}{
\begin{tabular}{llrrrr}
\toprule
Model & Predictor & Std. Coef. & 95\% CI & Robust $p$-value & VIF \\
\midrule
Span JEPA
& Target size & $0.121$ & $[0.120,\ 0.122]$ & $<0.001$ & $3.03$ \\
& Unique-code fraction & $-0.013$ & $[-0.015,\ -0.012]$ & $<0.001$ & $3.00$ \\
& Event-type diversity & $0.001$ & $[0.001,\ 0.002]$ & $<0.001$ & $1.56$ \\
& Target--context code overlap & $-0.014$ & $[-0.015,\ -0.013]$ & $<0.001$ & $2.28$ \\
& Code rarity & $-0.012$ & $[-0.012,\ -0.011]$ & $<0.001$ & $1.50$ \\
\midrule
Future JEPA
& Target size & $0.067$ & $[0.065,\ 0.068]$ & $<0.001$ & $3.32$ \\
& Unique-code fraction & $-0.069$ & $[-0.071,\ -0.068]$ & $<0.001$ & $2.91$ \\
& Event-type diversity & $-0.001$ & $[-0.002,\ 0.001]$ & $0.379$ & $1.86$ \\
& Target--context code overlap & $-0.021$ & $[-0.022,\ -0.020]$ & $<0.001$ & $1.96$ \\
& Code rarity & $-0.023$ & $[-0.025,\ -0.022]$ & $<0.001$ & $1.56$ \\
\midrule
Random JEPA
& Target size & $0.060$ & $[0.059,\ 0.062]$ & $<0.001$ & $6.37$ \\
& Unique-code fraction & $-0.040$ & $[-0.041,\ -0.039]$ & $<0.001$ & $6.56$ \\
& Event-type diversity & $-0.011$ & $[-0.011,\ -0.010]$ & $<0.001$ & $1.26$ \\
& Target--context code overlap & $-0.018$ & $[-0.018,\ -0.017]$ & $<0.001$ & $3.11$ \\
& Code rarity & $-0.009$ & $[-0.010,\ -0.008]$ & $<0.001$ & $1.20$ \\
\bottomrule
\end{tabular}}
\end{table*}

\subsection{Context and Target selection}
For span masking, let $N$ be the number of real (non-padding) events in the training window, and let $r$ be the mask ratio. The masked-token budget is $B=\lfloor N r\rfloor$, clipped to $[0,N]$. The number of spans is $4$ when $B\ge 40$, and $\max(1,\lfloor B/10\rfloor)$ otherwise. 
Each span has length has a rough length of $\lfloor B/S\rfloor$, where $S$ is the number of spans, and the minimum span length is 10 tokens. Its starting position is drawn independently and uniformly among the positions where a block of that length fits, so spans may overlap. Events that fall in no span are the context.

For future prediction, timestamps are measured in hours from the first event in the same window. The cut $T$ is drawn uniformly from event times that leave at least $10$ context events at times $\le T$ and at least $8$ target events in $(T,T+24\text{h}]$. Events after $T+24\text{h}$ are neither context nor target. If no such cut exists, the last $10\%$ of events in the window are the target and the preceding events are the context.

\subsection{Hyperparameters}
\label{Hyperparameters}
\begin{table}[H]
\centering
\caption{Hyperparameter sweep used during pretraining for the three JEPA objectives. Transformer and predictor settings were held fixed.}
\label{tab:hyperparameter_sweep}
\begin{tabular}{ll}
\toprule
Hyperparameter & Values \\
\midrule
Learning rate & $1\times10^{-5}$, $5\times10^{-5}$, $\mathbf{1\times10^{-4}}$, $5\times10^{-4}$ \\
Mask ratio & $\mathbf{0.15}$, $0.30$, $0.50$ \\
Batch size & $64$, $\mathbf{128}$, $256$ \\
EMA coefficient & \textbf{0.996} \\
\midrule
\multicolumn{2}{l}{\textit{Transformer (fixed)}} \\
Hidden size & \textbf{768} \\
Layers & \textbf{6} \\
Attention heads & \textbf{8} \\
Feed-forward width & \textbf{3072} \\
Dropout & \textbf{0.1} \\
Context length & \textbf{1024} \\
Normalization & pre-norm, final LayerNorm \\
Position encoding & RoPE \\
Attention & bidirectional \\
\midrule
\multicolumn{2}{l}{\textit{Predictor (fixed)}} \\
Hidden size & \textbf{768} \\
Layers & \textbf{2} \\
Attention heads & \textbf{8} \\
Feed-forward width & \textbf{3072} \\
Dropout & \textbf{0.1} \\
Position encoding & RoPE \\
Normalization & pre-norm, final LayerNorm \\
\bottomrule
\end{tabular}
\end{table}

EMA update:
$\theta_{\mathrm{tgt}} \leftarrow 0.996\,\theta_{\mathrm{tgt}} + 0.004\,\theta_{\mathrm{context}}$

\FloatBarrier
\subsection{MIMIC Dataset Information}
\begin{table}[H]
\centering
\caption{MIMIC-IV Population and Task-Specific Distribution}
\label{tab:mimic_combined}
\small
\begin{tabular}{lcccccc}
\toprule
& \multicolumn{2}{c}{\textbf{Train}} & \multicolumn{2}{c}{\textbf{Validation}} & \multicolumn{2}{c}{\textbf{Test}} \\
\cmidrule(lr){2-3} \cmidrule(lr){4-5} \cmidrule(lr){6-7}
\textbf{Metric / Task} & \textbf{Count} & \textbf{\%} & \textbf{Count} & \textbf{\%} & \textbf{Count} & \textbf{\%} \\
\midrule
\textit{Unique Patients} & 291,702 & 80.0\% & 36,463 & 10.0\% & 36,462 & 10.0\% \\
\midrule
\textbf{ICU Mortality} & & & & & & \\
\quad Case (Positive) & 5,881 & 7.8\% & 705 & 7.7\% & 715 & 7.7\% \\
\quad Control (Negative) & 69,169 & 92.2\% & 8,508 & 92.3\% & 8,628 & 92.3\% \\
\midrule
\textbf{ICU Readmission} & & & & & & \\
\quad Case (Positive) & 10,745 & 14.2\% & 1,345 & 14.5\% & 1,350 & 14.3\% \\
\quad Control (Negative) & 64,978 & 85.8\% & 7,947 & 85.5\% & 8,079 & 85.7\% \\
\midrule
\textbf{Hospital Mortality} & & & & & & \\
\quad Case (Positive) & 8,061 & 2.9\% & 1,014 & 2.9\% & 988 & 2.8\% \\
\quad Control (Negative) & 271,056 & 97.1\% & 34,271 & 97.1\% & 33,840 & 97.2\% \\
\midrule
\textbf{Hospital Readmission} & & & & & & \\
\quad Case (Positive) & 86,859 & 20.4\% & 11,074 & 20.6\% & 11,480 & 21.2\% \\
\quad Control (Negative) & 339,559 & 79.6\% & 42,659 & 79.4\% & 42,676 & 78.8\% \\
\bottomrule
\end{tabular}
\end{table}

\FloatBarrier
\subsection{Linear Probing}

\begin{table}[H]
\centering
\caption{
Downstream linear-probe performance on AUROC and AUPRC with brackets denoting 95\% percentile bootstrap confidence intervals from 1,000 stratified resamples
}
\label{tab:downstream_results}
\resizebox{\textwidth}{!}{
\begin{tabular}{lcccc}
\toprule
Model
& ICU Mortality
& ICU Readmission
& Hospital Mortality
& Hospital Readmission \\
\midrule

\multicolumn{5}{l}{\textbf{AUROC}} \\

MLM
& \makecell{\textbf{0.774} \\ {\scriptsize [0.757, 0.792]}}
& \makecell{\textbf{0.695} \\ {\scriptsize [0.677, 0.712]}}
& \makecell{\textbf{0.878} \\ {\scriptsize [0.867, 0.889]}}
& \makecell{\textbf{0.720} \\ {\scriptsize [0.711, 0.728]}} \\

Span JEPA
& \makecell{0.715 \\ {\scriptsize [0.696, 0.734]}}
& \makecell{0.642 \\ {\scriptsize [0.625, 0.658]}}
& \makecell{0.845 \\ {\scriptsize [0.832, 0.858]}}
& \makecell{0.700 \\ {\scriptsize [0.690, 0.711]}} \\

Future JEPA
& \makecell{0.757 \\ {\scriptsize [0.739, 0.775]}}
& \makecell{0.681 \\ {\scriptsize [0.663, 0.697]}}
& \makecell{0.870 \\ {\scriptsize [0.859, 0.882]}}
& \makecell{0.708 \\ {\scriptsize [0.698, 0.719]}} \\

Random JEPA
& \makecell{0.715 \\ {\scriptsize [0.696, 0.734]}}
& \makecell{0.638 \\ {\scriptsize [0.622, 0.653]}}
& \makecell{0.841 \\ {\scriptsize [0.827, 0.854]}}
& \makecell{0.691 \\ {\scriptsize [0.681, 0.702]}} \\

\midrule
\multicolumn{5}{l}{\textbf{AUPRC}} \\

MLM
& \makecell{\textbf{0.241} \\ {\scriptsize [0.214, 0.274]}}
& \makecell{\textbf{0.269} \\ {\scriptsize [0.246, 0.294]}}
& \makecell{\textbf{0.266} \\ {\scriptsize [0.231, 0.306]}}
& \makecell{\textbf{0.416} \\ {\scriptsize [0.398, 0.436]}} \\

Span JEPA
& \makecell{0.172 \\ {\scriptsize [0.154, 0.198]}}
& \makecell{0.219 \\ {\scriptsize [0.202, 0.238]}}
& \makecell{0.210 \\ {\scriptsize [0.178, 0.249]}}
& \makecell{0.401 \\ {\scriptsize [0.376, 0.428]}} \\

Future JEPA
& \makecell{0.216 \\ {\scriptsize [0.193, 0.247]}}
& \makecell{0.259 \\ {\scriptsize [0.238, 0.285]}}
& \makecell{0.249 \\ {\scriptsize [0.214, 0.289]}}
& \makecell{0.409 \\ {\scriptsize [0.383, 0.437]}} \\

Random JEPA
& \makecell{0.179 \\ {\scriptsize [0.159, 0.204]}}
& \makecell{0.215 \\ {\scriptsize [0.199, 0.232]}}
& \makecell{0.214 \\ {\scriptsize [0.182, 0.252]}}
& \makecell{0.390 \\ {\scriptsize [0.367, 0.416]}} \\

\bottomrule
\end{tabular}}
\end{table}

